\documentclass{wscpaperproc}
\usepackage{latexsym}
\usepackage{graphicx}
\usepackage{mathptmx}
\usepackage{tcolorbox}
\usepackage{enumitem}
\usepackage{statmath}
\usepackage[T1]{fontenc}

\usepackage{amsmath}
\usepackage{amsfonts}
\usepackage{amssymb}
\usepackage{amsbsy}
\usepackage{amsthm}
\usepackage{algpseudocode}
\usepackage{algorithm}

\usepackage[pdftex,colorlinks=true,urlcolor=blue,citecolor=black,anchorcolor=black,linkcolor=black]{hyperref}

\newtheoremstyle{wsc}
{3pt}
{3pt}
{}
{}
{\bf}
{}
{.5em}
{}
\newcommand{\indep}{\perp \!\!\! \perp}
\newcommand{\notindep}{\not\!\perp\!\!\!\perp}
\theoremstyle{wsc}
\newtheorem{theorem}{Theorem}
\newtheorem{assumption}{Assumption}
\newtheorem{proposition}{Proposition}
\newtheorem{remark}{Remark}

\newtheorem{definition}{Definition}

\newcommand{\algorithmicinput}{\textbf{input}}
\newcommand{\algorithmicoutput}{\textbf{output}}
\newcommand{\algorithmicinit}{\textbf{First stage: model $P\left(Z=1|\mathbf{s}\right)$\hspace{0.05cm}}}
\newcommand{\algorithmicquery}{\textbf{Second stage: bimodal query\hspace{0.05cm}}}
\newcommand{\algorithmictest}{\textbf{Third stage: FR two-sample test\hspace{0.05cm}}}
\begin{document}

%
%

\pagestyle{fancyplain}

\thispagestyle{plain}
\firstPageHead{}

\chead{\fancyplain{}{\itshape Li, Berisha, and Dasarathy }}

\rhead{}
\cfoot{}
\renewcommand{\headrulewidth}{0pt} 

\makeatletter
\let\@internalcite\cite
\def\cite{\def\@citeseppen{-1000}%
    \def\@cite##1##2{(##1\if@tempswa , ##2\fi)}%
    \def\citeauthoryear##1##2##3{##1 ##3}\@internalcite}
\def\citeNP{\def\@citeseppen{-1000}%
    \def\@cite##1##2{##1\if@tempswa , ##2\fi}%
    \def\citeauthoryear##1##2##3{##1 ##3}\@internalcite}
\def\citeN{\def\@citeseppen{-1000}%
    \def\@cite##1##2{##1\if@tempswa, ##2)\else{}\fi}%
    \def\citeauthoryear##1##2##3{##1 (##3)}\@citedata}
\def\citeA{\def\@citeseppen{-1000}%
    \def\@cite##1##2{(##1\if@tempswa , ##2\fi)}%
    \def\citeauthoryear##1##2##3{##1}\@internalcite}
\def\citeANP{\def\@citeseppen{-1000}%
    \def\@cite##1##2{##1\if@tempswa , ##2\fi}%
    \def\citeauthoryear##1##2##3{##1}\@internalcite}
\def\shortcite{\def\@citeseppen{-1000}%
    \def\@cite##1##2{(##1\if@tempswa , ##2\fi)}%
    \def\citeauthoryear##1##2##3{##2 ##3}\@internalcite}
\def\shortciteNP{\def\@citeseppen{-1000}%
    \def\@cite##1##2{##1\if@tempswa , ##2\fi}%
    \def\citeauthoryear##1##2##3{##2 ##3}\@internalcite}
\def\shortciteN{\def\@citeseppen{-1000}%
    \def\@cite##1##2{##1\if@tempswa, ##2\else{}\fi}%
    \def\citeauthoryear##1##2##3{##2 (##3)}\@citedata}
\def\shortciteA{\def\@citeseppen{-1000}%
    \def\@cite##1##2{(##1\if@tempswa , ##2\fi)}%
    \def\citeauthoryear##1##2##3{##2}\@internalcite}
\def\shortciteANP{\def\@citeseppen{-1000}%
    \def\@cite##1##2{##1\if@tempswa , ##2\fi}%
    \def\citeauthoryear##1##2##3{##2}\@internalcite}
\def\citeyear{\def\@citeseppen{-1000}%
    \def\@cite##1##2{(##1\if@tempswa , ##2\fi)}%
    \def\citeauthoryear##1##2##3{##3}\@citedata}
\def\citeyearNP{\def\@citeseppen{-1000}%
    \def\@cite##1##2{##1\if@tempswa , ##2\fi}%
    \def\citeauthoryear##1##2##3{##3}\@citedata}
%
%
%
\def\@citedata{%
    \@ifnextchar [{\@tempswatrue\@citedatax}%
                  {\@tempswafalse\@citedatax[]}%
}

\def\@citedatax[#1]#2{%
\if@filesw\immediate\write\@auxout{\string\citation{#2}}\fi%
  \def\@citea{}\@cite{\@for\@citeb:=#2\do%
    {\@citea\def\@citea{, }\@ifundefined
       {b@\@citeb}{{\bf ?}%
       \@warning{Citation `\@citeb' on page \thepage \space undefined}}%
{\csname b@\@citeb\endcsname}}}{#1}}%

%
\def\@citex[#1]#2{%
\if@filesw\immediate\write\@auxout{\string\citation{#2}}\fi%
  \def\@citea{}\@cite{\@for\@citeb:=#2\do%
    {\@citea\def\@citea{; }\@ifundefined
       {b@\@citeb}{{\bf ?}%
       \@warning{Citation `\@citeb' on page \thepage \space undefined}}%
{\csname b@\@citeb\endcsname}}}{#1}}%

%
\def\@biblabel#1{}
\makeatother



\newdimen\bibindent
\bibindent=0.0em
\def\thebibliography#1{\section*{\refname}\list
   {}{\settowidth\labelwidth{[#1]}
   \leftmargin\parindent
   \itemindent -\parindent
   \listparindent \itemindent
   \itemsep 0pt
   \parsep 0pt}
   \def\newblock{}
   \sloppy
   \sfcode`\.=1000\relax}


\setlength{\baselineskip}{12.7pt}

\title{ADVANCED TUTORIAL: LABEL-EFFICIENT TWO-SAMPLE TESTS}

\author{\begin{center}Weizhi Li\textsuperscript{1,}\textsuperscript{2}, Visar Berisha\textsuperscript{2},  and Gautam Dasarathy\textsuperscript{2}\\
[11pt]
\textsuperscript{1}Los Alamos National Laboratory, Los Alamos, NM, USA\\
\textsuperscript{2}Arizona State University, Tempe, AZ, USA\end{center}
}

\maketitle

\vspace{-12pt}
\section*{ABSTRACT}

Hypothesis testing is a statistical inference approach used to determine whether data supports a specific hypothesis. An important type is the two-sample test, which evaluates whether two sets of data points are from identical distributions. This test is widely used, such as by clinical researchers comparing treatment effectiveness. This tutorial explores two-sample testing in a context where an analyst has many features from two samples, but determining the sample membership (or labels) of these features is costly. In machine learning, a similar scenario is studied in active learning. This tutorial extends active learning concepts to two-sample testing within this \textit{label-costly} setting while maintaining statistical validity and high testing power. Additionally, the tutorial discusses practical applications of these label-efficient two-sample tests.
\section{Introduction}
\label{sec:intro}
Much scientific research involves conducting experiments and drawing conclusions from the results. This process is prevalent in fields such as physics, chemistry, biology, and other natural sciences, where scientists perform extensive experiments to verify hypotheses. 
To address the common pitfall of scientists making false discoveries or incorrectly claiming a hypothesis is true, statisticians have developed hypothesis testing, a statistical inference approach that evaluates whether a hypothesis is supported by experimental data.

Conducting experiments is often resource-intensive. For example, clinical trials in drug discovery can be very costly, running an emulator for physics experiments can demand significant computational power, and chemical experiments may involve lengthy reaction times. To address these challenges, the research community has proposed ``data-efficient" hypothesis testing~\shortcite{kartik2019active,naghshvar2013active,cohen2015active,naghshvar2010active,dasarathy2017sketched,cecchi2017adaptive}, which aims to perform a limited number of experiments to make effective decisions. In this approach, an action space that characterizes the data sources (e.g., various experiments) is typically defined. The hypothesis testing procedure then adaptively selects an informative data source to facilitate decision-making regarding the hypotheses. This tutorial is focused on an important type of hypothesis testing, \textit{two-sample hypothesis tests}, in a \textit{label-costly} setting.\\
\textbf{Two-sample testing}: Two-sample tests are applied to data samples (or measurements) from two distributions to determine if the data support the hypothesis that the distributions are different. If we consider each data point as a feature and label (indicating which distribution the data is from) pair, then two-sample testing is equivalent to testing the dependence between the features and the labels. Viewing the problem through this lens, the null hypothesis for two-sample testing states that the feature and label variables are independent, while the alternative hypothesis states the opposite. An analyst performing a two-sample test is to decide between the null and alternative hypotheses based on the data from the two distributions. Depending on the flexibility in examining the two samples, she can choose to perform a \textit{batch} two-sample test, where hypothesis testing is conducted only after all data collection is complete, with no continuous data collection and testing allowed. Alternatively, she can opt for \textit{sequential} two-sample testing, where both data collection and hypothesis testing proceed sequentially and concurrently.\\
\textbf{Label-costly setting}: In traditional two-sample testing, the underlying assumption is that both the features and their corresponding labels are simultaneously available. 
In this tutorial, we discuss two-sample testing in a rarely considered setting, where the measurements (or features) are readily accessible, but their groups (or labels) are unknown and difficult/costly to obtain. This scenario is analogous to active learning~\shortcite{cohn1994improving} in machine learning community, leading to many label-efficient algorithms~\shortcite{gal2017deep,balcan2006agnostic,dasarathy2015s2,hanneke2014theory} for classification purposes. In contrast, this tutorial focuses on making decisions on hypotheses, i.e., hypothesis testing, which requires a different label query strategy to achieve label efficiency while effectively determining whether the null hypothesis is supported by the data.\\\looseness=-1
\textbf{Motivating application}: 
Computer scientists and clinicians have collaborated to develop digital twins for simulating patient data in clinical trials. Identifying potential biomarkers indicative of disease severity is crucial before constructing these twins. Without this step, significant resources could be wasted on non-indicative biomarkers. Clinicians must validate these biomarkers through experiments, which can be costly. Efficient validation against expensive and labor-intensive labels is essential. Label-efficient two-sample tests address this challenge by enabling validation with a small subset of labeled data. A motivating example is the validation of digital biomarkers in Alzheimer’s disease relative to imaging markers. Say we want to determine whether a series of digital biomarkers that can be collected at scale (e.g. gait, speech, typing speed measured using a patient’s smartphone) is related to amyloid buildup in the brain (measured via neuroimaging, and an indication of increased risk of Alzheimer’s disease). In this scenario, we can obtain the digital biomarkers on a large scale by distributing the tests via the internet. However, actually determining if a particular patient is amyloid positive (higher risk of Alzheimer’s disease) or negative (lower risk) involves bringing participants in expensive and burdensome neuroimaging tests. It is of considerable interest to validate the digital biomarkers relative to the biomarker by only collecting neuroimaging data from a small number of participants. Notice that this scenario is in stark contrast to traditional formulations of two sample testing, where the class label (amyloid positivity) is assumed to be available for participants in the study.\\
\textbf{Questions to address}: Developing a legitimate label-efficient two-sample test requires addressing the following questions,
\begin{itemize}[topsep=0.1pt, partopsep=0.1pt, itemsep=0pt, parsep=0pt]
\item What is an effective query scheme for the label querying?
\item Does the developed test maintain statistical validity, i.e., can the error of rejecting the null hypothesis be controlled when the null is true?
\item Does the developed test have high testing power, i.e., can it reject the null hypothesis with high probability when the alternative hypothesis is true?
\end{itemize}
In this tutorial, we discuss label-efficient two-sample testing and demonstrate that it is possible to achieve performance similar to traditional two-sample testing while minimizing the number of label queries. We will present label-efficient testing algorithms that can answer these questions with supporting theorems. The organization of this tutorial is as follows. In Section~\ref{sec:traditionaltest}, we review the two-sample test.  In section~\ref{sec:activetest}, we present the label-efficient two-sample testing problem. In Section~\ref{Sec:Bimodal-query}, we introduce an effective label query scheme called the bimodal query. In Section~\ref{sec:batch_activetest} and~\ref{sec:sequential_activetest}, we present batch and sequential label-efficient two-sample tests, respectively.
\section{Review of the Two-Sample Testing}
\label{sec:traditionaltest}
\subsection{A Traditional Two-Sample Testing Problem}
\label{subsec:traditionaltest}
Two-sample hypothesis testing evaluates whether two samples (or sets of data points)  are generated from the same distribution (null hypothesis) or different distributions (alternative hypothesis). Let $\mathcal{X} = \left\{\mathbf{x}_1, \ldots \mathbf{x}_{n_0}\right\}$ and $\mathcal{Y} = \left\{\mathbf{y}_1, \ldots, \mathbf{y}_{n_1}\right\}$ denote two samples of the realizations of \textit{i.i.d.} random variables  $\{\mathbf{X}_i\}_{i=1}^{n_0}$ and  $\{\mathbf{Y}_i\}_{i=1}^{n_0}$ for $\mathbf{X}\sim p_{\mathbf{X}}\left(\mathbf{x}\right)$ and $\mathbf{Y}\sim p_{\mathbf{Y}}\left(\mathbf{y}\right)$. The null and alternative hypotheses are formulated as $H_0$ and $H_1$ in the following:\looseness=-1
\begin{align}
    &H_0 :  p_{\mathbf{X}}\left(\mathbf{x}\right)=p_{\mathbf{Y}}\left(\mathbf{y}\right), \nonumber\\ 
    &H_1: p_{\mathbf{X}}\left(\mathbf{x}\right)\neq p_{\mathbf{Y}}\left(\mathbf{y}\right) .
    \label{eq:nullandalternative}
\end{align}
Then, \textbf{the \textit{traditional} two-sample testing problem is to decide whether to reject or retain $H_0$ after examining $\mathcal{X}$ and $\mathcal{Y}$}.
\subsection{Desired Properties for the Traditional Two-Sample Testing}
\label{subsec:DesiredProp}
A conventional two-sample test is formulated as follows~\shortciteNP{johnson2011elementary}: (a) the analyst obtains two sets of data points $\mathcal{X} = \left\{\mathbf{x}_1 ,\ldots, \mathbf{x}_{n_0}\right\}$ and $\mathcal{Y}=\left\{\mathbf{y}_1, \ldots, \mathbf{y}_{n_1}\right\}$; (b) she computes a test statistic $t$ from $\mathcal{X}$ and $\mathcal{Y}$; (c) she then computes the $p$-value for the statistic $t$ under $H_0$ (both $\mathcal{X}$ and $\mathcal{Y}$ come from the same distribution). Formally, the $p$-value of a \textit{valid} two-sample test represents the probability of obtaining test results at least as extreme as the test statistic $t$, assuming the null $H_0$ is true. In other words, a low $p$-value indicates that the test statistic $t$ is unlikely to have been generated under $H_0$, suggesting that $H_0$ should be rejected. Typically, the analyst compares the $p$-value to a predefined significance-level $\alpha$ and reject $H_0$ if $p\leq\alpha$. Throughout this tutorial, we will consider the following desired properties for a two-sample test:\\
\textbf{Valid $p$-value}: Given $p_{\mathbf{X}}\left(\mathbf{x}\right)$, $p_{\mathbf{Y}}\left(\mathbf{y}\right)$, the resulting realization sets $\mathcal{X}$ and $\mathcal{Y}$  and a significance level $\alpha$, a two-sample test generates a \textit{valid} $p$-value computed from $\mathcal{X}\bigcup\mathcal{Y}$ if $P\left(p\leq\alpha\right)\leq\alpha$ when $H_0$ is true. \\
\textbf{Consistency}: Given $p_{\mathbf{X}}\left(\mathbf{x}\right)$, $p_{\mathbf{Y}}\left(\mathbf{y}\right)$, sets $\mathcal{X}$ and $\mathcal{Y}$ containing $n$ total samples drawn from these distributions, and a significance level $\alpha$, a two-sample test is \textit{consistent} if $\lim_{n\to\infty}P\left(p\leq\alpha\right)=1$ when $H_1$ is true.\\
\textbf{High testing power}: Given $p_{\mathbf{X}}\left(\mathbf{x}\right)$, $p_{\mathbf{Y}}\left(\mathbf{y}\right)$, $n=|\mathcal{X}| + |\mathcal{Y}|$, the resulting realization sets $\mathcal{X}$ and $\mathcal{Y}$ and a significance level $\alpha$, a two-sample test has high testing power if $P\left(p\leq\alpha\right)$ is high for the finite $n$ when $H_1$ is true. In other words, a two-sample test is more powerful than another if its $P\left(p\leq\alpha\right)$ is higher given a specific $n$.\looseness=-1

Here, we write $P_0(p \leq \alpha)$ and $P_1(p \leq \alpha)$ to denote $P(p \leq \alpha)$ under $H_0$ and $H_1$, respectively. $P_0(p \leq \alpha)$ represents the \textit{Type I error}, which is the probability of making a wrong decision when $H_0$ is true, while $1 - P_1(p \leq \alpha)$ represents the \textit{Type II error}, which is the probability of making a wrong decision when $H_1$ is true. A valid two-sample test (i.e., one that produces a valid $p$-value) can upper-bound its Type I error $P_0$ by a predefined significance level $\alpha$. On the other hand, the properties of consistency and high testing power serve to reduce the Type II error $1 - P_1$.
\subsection{Classical Two-Sample Tests}
The development of two-sample tests has a rich history~\shortcite{friedman1979multivariate,chen2017new,hotelling1992generalization,friedman2004multivariate,vayatis2009auc,lheritier2018sequential,hajnal1961two}. In this section, we introduce representative examples of the non-parametric two-sample test and the sequential non-parametric two-sample test. In section \ref{sec:activetest}, we will see how label-efficient two-sample tests are extended from these classical two-sample tests.

\subsubsection{Nonparametric Two-Sample Testing}
\label{subsubsec:FRTest}
This section introduces the Friedman-Rafsky (FR) test~\shortcite{friedman1979multivariate}, a non-parametric, graph-based test to determine whether two multivariate samples are realization of the same distribution. The null and alternative hypotheses and the notations used are stated in~\eqref{eq:nullandalternative}.\\ 
\textbf{The FR test statistic}: The FR test statistic is computed as follows. First, one constructs a Euclidean minimum spanning tree (MST) over the samples $\mathcal{X}\bigcup \mathcal{Y}$, i.e., the MST of a complete graph whose vertices are the samples, and edge weights are the Euclidean distance between the samples. Then, one counts the edges connecting samples from opposite classes (i.e., cut edges).  We use $r_{n}$ to denote the cut-edge number for the MST constructed over two samples $\mathcal{X}\bigcup \mathcal{Y}$ with a total size of $n$, and use $R_n$ to denote the corresponding random variable.
Under the alternative $H_1$, $r_{n}$ is expected to be small, and under the null  $H_0$, $r_{n}$ is expected to be large. The FR test statistic $w_n$ is a normalized version of $r_n$,
\begin{align}
w_n = \frac{r_n-\mathbb{E}\left[R_n\mid H_0\right]}{\sqrt{{\rm Var}\left[R_n\mid H_0\right]}},
\label{FRstat}
\end{align}
where $\mathbb{E}\left[R_n\mid H_0\right]$ and ${\rm Var}\left[R_n\mid H_0\right]$ are the expectation and the variance of $R_{n}$ under the null $H_0$.
We use $W_n$ to denote a random variable of which $w_n$ is a realization.  The FR test rejects $H_0$ if a small $W_n$ is observed.\\ 
\textbf{A permutation test}: In practice, as stated in~\shortciteNP{friedman1979multivariate}, the FR test is carried out as a permutation test where the null distribution (distribution of a statistic under the null $H_0$) of $W_n$ is obtained by calculating all possible values of $w_n$~\eqref{FRstat} under all possible rearrangements of the observations of $\mathcal{X}\bigcup\mathcal{Y}$. Then a $p$-value is obtained using the permutation null distribution and the $w_n$ computed from $\mathcal{X}$ and $\mathcal{Y}$. The  $p$-value is compared to a  significance level $\alpha$ to reject $H_0$ for $p\leq\alpha$. We refer readers to~\shortciteNP{welch1990construction} for the procedure of the permutation test.\\ 
\textbf{Approximation of the permutation test}: Both Theorem 4.1.2 in~\shortciteNP{bloemena1964sampling} and Section 4 in~\shortciteNP{friedman1979multivariate} demonstrate that, if $W_n$ is generated under $H_0$, then the \emph{permutation distribution} of $W_n$ approaches a standard normal distribution for large sample size $n\to\infty$: $ W_n\xrightarrow{\mathcal{D}}\mathcal{N}(0,1)$, where $\xrightarrow{\mathcal{D}}$ stands for distributional convergence. Therefore, a valid $p$-value for the FR test is given by 
\begin{align}
   p = \Phi[W_n],  
  \label{Pvalue}
\end{align}
An analyst can construct the FR statistic $w_n$ to generate the $p$-value and compare the $p$-value with $\alpha$ to decide $H_0$ or $H_1$. The Type I error is upper-bounded by $\alpha$.\\
\textbf{Consistency}: The consistency of the FR test is proved in~\shortciteNP{henze1999multivariate} based on the following.
\begin{theorem}{\shortcite{henze1999multivariate}}
Given $\mathcal{X}=\left\{\mathbf{x}_1,\cdots,\mathbf{x}_{n_0}\right\}$ and $\mathcal{Y}=\left\{\mathbf{y}_1,\cdots,\mathbf{y}_{n_1}\right\}$ which are \textit{i.i.d.} realizations of $\mathbf{X}\sim p_{\mathbf{X}}\left(\mathbf{x}\right)$ and $\mathbf{Y}\sim p_{\mathbf{Y}}\left(\mathbf{y}\right)$, and  $n=n_0+n_1$, suppose $\lim_{n\to\infty}\frac{n_0}{n}=u$, $\lim_{n\to\infty}\frac{n_1}{n}=v$ where $u+v=1$, then 
\begin{align}
    \frac{R_n}{n}\inas 2uv\int\frac{p_{\mathbf{X}}(\mathbf{s})p_{\mathbf{Y}}(\mathbf{s})}{up_\mathbf{X}(\mathbf{s}) + vp_\mathbf{Y}(\mathbf{s})}d\mathbf{s}.
    \label{FRAsymp}
\end{align}
\label{TheoremHenze}
\end{theorem}
Under $H_0$, where $p_\mathbf{X}\left(\mathbf{s}\right)=p_\mathbf{Y}\left(\mathbf{s}\right)$, the R.H.S of~\eqref{FRAsymp} is strictly larger than that under $H_1$, where  $p_\mathbf{X}\left(\mathbf{s}\right)\neq p_\mathbf{Y}\left(\mathbf{s}\right)$, resulting in the consistency of the FR test. 
 \subsubsection{Sequential Nonparametric Two-Sample Testing}
 \label{subsubsec:SequentialStatistic}
FR~\shortcite{friedman1979multivariate} is a \textit{batch} two-sample test, where two samples, $\mathcal{X}$ and $\mathcal{Y}$ are collected, and then a two-sample hypothesis test is performed on $\mathcal{X}\bigcup\mathcal{Y}$ to decide on $H_0$ or $H_1$. In this method,  \textit{continuous data collection and hypothesis testing are not allowed}. The research community has proposed sequential two-sample tests~\shortcite{wald1992sequential,lheritier2018sequential,hajnal1961two,shekhar2021game,balsubramani2015sequential} that allow the analyst to sequentially collect data and monitor statistical evidence, i.e., a statistic computed from the data. The test can stop anytime when sufficient evidence has been accumulated to make a decision.  

 In this section, we present a \textit{sequential nonparametric two-sample test} constructed by~\shortciteNP{lheritier2018sequential}.  We consider the observed data  a \textit{measurement sequence}  $\mathcal{S}=\mathcal{X}\bigcup\mathcal{Y}=\left\{\mathbf{s}_i\right\}_{i=1}^n$ and a corresponding \textit{label sequence} $\mathcal{Z}=\left\{z_1, \ldots, z_n\right\}$, where $z_i = 0$ if $\mathbf{s}_i \in \mathcal{X}$ and 1 otherwise. Accordingly, our observation model is $n$ \textit{i.i.d.} draws from the joint distribution $p_{\mathbf{S}Z}\left(\mathbf{s}, z\right)$. The two sample testing problem under this formulation is equivalent to testing if $p_{\mathbf{S}\mid Z}\left(\cdot\mid 0\right) = p_{\mathbf{S}\mid Z}\left(\cdot\mid 1\right)$ (i.e., $\mathbf{S}$ and $Z$ are independent).\\
 \textbf{\shortcite{lheritier2018sequential} test statistic}:~\shortciteNP{lheritier2018sequential} assumes the class prior $P_Z(z)$ is known in the problem setting. Additionally, they build a \textit{nonparametric} class-probability predictor $Q\left(z\mid \mathbf{s}\right)$ to model $P_{Z\mid\mathbf{S}}\left(z\mid \mathbf{s}\right)$. Specifically, $Q_n\left(z\mid \mathbf{s}\right)$ represents a class-probability predictor built with the past observed sample sequence $\left\{\mathbf{s}_i, z_i\right\}^{n-1}_{i=1}$ to model $P_{Z\mid \mathbf{S}}\left(z_n \mid \mathbf{s}_n\right)$--the posterior probability of $z_n$ given newly observed $\mathbf{s}_n$. Any nonparametric probabilistic classifier, such as a decision tree  and KNN classifier, can be used to build $Q_{n}(z\mid \mathbf{s})$.  The following statistic is then constructed,
 \begin{align}
    t_n=\frac{P_Z(z^n)}{Q(z^n\mid \mathbf{s}^n)}=\prod_{i=1}^n\frac{P_Z(z_i)}{Q_{i}(z_i\mid \mathbf{s}_i)}.
\label{ratiostatisics}
\end{align}
The statistic~\eqref{ratiostatisics} stems from the well-known likelihood ratio test~\shortcite{wilks1938large}. Here, the likelihood under $H_0$ where $\mathbf{S} \indep Z$ is $\prod_{i=1}^np_\mathbf{S}\left(\mathbf{s}_i\right)P_Z\left(z_i\right)$ and the likelihood under $H_1$ where $\mathbf{S} \notindep Z$ is $\prod_{i=1}^nP_{Z\mid\mathbf{S}}\left(z_i\mid\mathbf{s}_i\right)p_\mathbf{S}\left(s_i\right)$. Replacing $P_{Z\mid\mathbf{S}}$ with an approximation $Q$ in the likelihood ratio $\prod_{i=1}^n\frac{p_\mathbf{S}\left(\mathbf{s}_i\right)P_Z\left(z_i\right)}{P_{Z\mid\mathbf{S}}\left(z_i\mid\mathbf{s}_i\right)p_\mathbf{S}\left(s_i\right)}$ leads to the~\shortcite{lheritier2018sequential} statistic in~\eqref{ratiostatisics}.\\
\textbf{Anytime-valid $p$-value}: Typically, a sequential test iteratively collects $\left(\mathbf{s}, z\right)$, adds $\left(\mathbf{s}, z\right)$ to the measurement and label sequences $\mathcal{S}$ and $\mathcal{Z}$, and monitors the $p$-values computed from the sequences for every $n$. Implementing the FR in a sequential way generates \textit{invalid} $p$-values, meaning that $P\left(p\leq\alpha\right)$ is not necessarily upper-bounded by $\alpha$ for every $n$. This occurs when multiple $p$-values are generated without \textit{adjustment} for deciding between $H_0$ and $H_1$, the probability of observing a rare event increases, thereby raising the likelihood of incorrectly rejecting $H_0$ (i.e., making a Type I error); alternatively, one may make the test extremely weak by using a very conservative Bonferroni correction (union bound)~\cite{dunn1961Multiple} across all time steps. In contrast, the~\shortcite{lheritier2018sequential} statistic in~\eqref{ratiostatisics} produces the \textit{anytime-valid} $p$-value stated in the following theorem,
\begin{theorem}\shortcite{lheritier2018sequential}
 An analyst sequentially collects $\left(\mathbf{s}, z\right)$ resulting in $\left\{\mathbf{s}_i,z_i\right\}_{i=1}^n$ which are \textit{i.i.d.} realizations of $(\mathbf{S}, Z)\sim p_{\mathbf{S}Z}\left(\mathbf{s},z\right)$, and computes the statistic $t_n$ as in~\eqref{ratiostatisics} with $\left\{\mathbf{s}_i,z_i\right\}_{i=1}^n$. Under $H_0$, the following holds for the random variable $T_n$ from which $t_n$ is realized, 
\begin{align}
P_0\left(\exists n\geq1, T_{n}=\prod_{i=1}^{n}\frac{P(Z_i)}{Q_{i}\left(Z_i\mid \mathbf{S}_i\right)}\leq\alpha\right)\leq\alpha.
\end{align}
\label{PassiveTypeI}
\end{theorem}
Theorem~\ref{PassiveTypeI} implies that $t_n$ is an anytime-valid $p$-value and rejects $H_0$ whenever $t_n\leq\alpha$ leading to a Type I error upper-bounded by $\alpha$. Briefly, it is proved by observing that the sequence $\left(\frac{1}{T_1},\cdots,\frac{1}{T_n}\right)$ is a non-negative martingale, and hence Ville's maximal inequality~\shortcite{durrett2019probability,doob1939jean} can be used to bound the tail probability of the infimum of an infinite $\{T_i\}_{i=1}^n$ to develop Theorem~\ref{PassiveTypeI}.\\
\textbf{Consistency}: The authors of~\shortciteNP{gyorfi2002distribution} prove that when $Q\left(z\mid \mathbf{s}\right)$ is a kernel, KNN or partition estimates with proper smoothing parameters (e.g., bandwidth for the kernel), then $Q_n\left(z\mid \mathbf{s}\right)$ converges to $P\left(z\mid \mathbf{s}\right)$. Then, the following theorem holds:
\begin{theorem}
    When $Q\left(z\mid \mathbf{s}\right)$ is a kernel, KNN or partition estimate with proper smoothing parameters (e.g., bandwidth for the kernel), then
    \begin{align}
        \lim_{n\to\infty}\frac{\log T_n }{n}= \lim_{n\to\infty}\frac{1}{n}\sum_{i=1}^n\log\frac{P_Z(Z_i)}{Q_i(Z_i\mid\mathbf{S}_i)}=-\left(H\left(Z\right)-H\left(Z\mid\mathbf{S}\right)\right)=-I\left(\mathbf{S},Z\right)
    \end{align}
where $H(Z)$, $H\left(Z\mid\mathbf{S}\right)$, and $I\left(\mathbf{S};Z\right)$ are the entropy, conditional entropy and mutual information for $\left(\mathbf{S}, Z\right)\sim p_{\mathbf{S}Z}\left(\mathbf{s}, z\right)$.
\label{ConsistencyMI}
\end{theorem}
Theorem~\ref{ConsistencyMI} implies that when $Q$ is built with a proper nonparametric method, $\frac{\log T_n }{n}$ converges to the negation of mutual information, which is smaller than zero under $H_1$ where $\mathbf{S}\notindep Z$, leading to $\lim_{n\to\infty}P_1\left(T_n\leq\alpha\right)=1$. 

\section{The Label-Efficient Two-Sample Testing Problem} 
\label{sec:activetest}
To motivate our \textit{label-efficient} two-sample testing problem, we use a pair of random variables  $\left(\mathbf{S}, Z\right)$ to denote a feature and its label variables whose realization is $\left(\mathbf{s}, z\right)\in\mathbb{R}^d\times \{0,1\}$. The variable pair $\left(\mathbf{S}, Z\right)$ admits a joint distribution $p_{\mathbf{S}Z}(\mathbf{s},z)$. Furthermore, we write $\mathbb{S}$ to denote the support of $p_{\mathbf{S}}(\mathbf{s})$. 

In the traditional two-sample testing problem, discussed in section~\ref{subsec:traditionaltest}, it is assumed that both features and their corresponding labels are available simultaneously. This tutorial introduces a novel problem formulation where an analyst has free access to an unlabeled feature set $\mathcal{S}_u=\{\mathbf{s}_i\}_{i=1}^n$, but obtaining the corresponding labels $z_i$ in the label set $\mathcal{Z}$ is costly. The analyst is provided with a label budget $N_q \leq n$, allowing her to select a subset $\overline{\mathcal{S}} \subseteq \mathcal{S}_u$ with at most $N_q$ features, for which an oracle returns the corresponding label set $\overline{\mathcal{Z}} \subseteq \mathcal{Z}$. The original unlabeled feature set $\mathcal{S}_u$ consists of realizations of \textit{i.i.d.} variables $\mathbf{S}\sim p_{\mathbf{S}}\left(\mathbf{s}\right)$, but the selected $\overline{\mathcal{S}}$ \textit{does not} necessarily contain \textit{i.i.d.} variables. 

The label-efficient two sample testing problem under this formulation aims to test the following null and alternative hypotheses:
\begin{align}
    &H_0 :  p_{S|Z}(\cdot\mid 0) = p_{S|Z}(\cdot\mid 1) \qquad 
    H_1: p_{S|Z}(\cdot\mid 0)\neq p_{S|Z}(\cdot\mid 1)\nonumber,
\end{align}
or equivalently, to test the independence between $\mathbf{S}$ and $Z$:
\begin{align}
    H_0 :  p_{\mathbf{S}Z}(\mathbf{s},z)=p_\mathbf{S}(\mathbf{s})P_Z(z),\forall \mathbf{s}\in\mathbb{S} \qquad 
    H_1: p_{\mathbf{S}Z}(\mathbf{s},z)\neq p_\mathbf{S}(\mathbf{s})P_Z(z), \exists\mathbf{s}\in\mathbb{S}\nonumber
\end{align}
by examining $\overline{\mathcal{S}}$ and $\overline{\mathcal{Z}}$, whose sizes do not exceed the label budget $N_q$. Moving forward, we omit the subscripts in $p_{\mathbf{S}Z}(\mathbf{s},z)$, $P_Z(z)$ and $p_\mathbf{S}(\mathbf{s})$ and write them as $p(\mathbf{s},z)$, $P(z)$ and $p(\mathbf{s})$.  In addition, we use $\mathbf{s}^N$, $z^N$ and $(\mathbf{s},z)^N$ to denote sequences of samples $\{\mathbf{s}_i\}_{i=1}^N$, $\{z_i\}_{i=1}^N$ and $\{(\mathbf{s},z)_i\}^N_{i=1}$ respectively.
\section{Bimodal query}
\label{Sec:Bimodal-query}
Our label-efficient tests employ a probabilistic classification-based query scheme called the \textit{bimodal query} to determine which features to select from an unlabeled feature set. This query scheme was first proposed, to the best of our knowledge, in \shortciteNP{li2022label}.
Let $\overline{\mathcal{S}}$ and $\overline{\mathcal{Z}}$ be the  features and their revealed labels. The bimodal query uses a class-probability predictor $Q(z \mid \mathbf{s})$ and the unlabeled set $\mathcal{S}_u$ as input. This classifier $Q(z \mid \mathbf{s})$ is trained using $\overline{\mathcal{S}}$ and $\overline{\mathcal{Z}}$. The bimodal query then selects two features, $s_{q_0}\in\mathcal{S}_u$ and $s_{q_1}\in\mathcal{S}_u$ to query for their labels, where $s_{q_0}=\arg\max_{\mathbf{s}}\left[Q(Z=0|\mathbf{s})\right]$ and $s_{q_1}=\arg\max_{\mathbf{s}}\left[Q(Z=1|\mathbf{s})\right],\forall \mathbf{s}\in\mathcal{S}_u$. In other words, the features predicted to have the highest class one or zero posterior probabilities are labeled, as these features lie in the most "informative" region. 

The authors of~\shortciteNP{gyorfi2002distribution}  prove that when $Q\left(z\mid \mathbf{s}\right)$ is a kernel, KNN or partition estimates with proper smoothing parameters (e.g., bandwidth for the kernel) and labels are sufficiently revealed in the proximity of $\mathbf{s},\forall \mathbf{s}\in\mathbb{S}$, then $Q\left(z\mid \mathbf{s}\right)$ converges to $P\left(z\mid \mathbf{s}\right)$. A bimodal query using $Q$ with such a convergence property leads to the following,
\begin{definition}(\emph{Consistent bimodal query})
Let $\mathbb{S}$ be the support of  $p(\mathbf{s})$ and let $\mathcal{S}_u$ be an unlabeled set where sample features are  sampled \textit{i.i.d.} from $p(\mathbf{s})$, and let $P(z\mid \mathbf{s})$ denote the posterior probability of $z$ given $\mathbf{s}\in\mathbb{S}$.  An analyst adopts a label query scheme, for every $n>0$,  to query the label $Z_n$ of $\mathbf{S}_n\in\mathcal{S}_u$ such that $\mathbf{S}_n$ admits a probability density function (pdf) $p_n(\mathbf{s})$. The label query scheme is a \textit{consistent bimodal query} if $\lim_{n\to\infty}p_n\left(\mathbf{s}\right)=p^*\left(\mathbf{s}\right)$ where
 \begin{align}
     p^*(\mathbf{s})&=0,\forall \mathbf{s}\in\mathbb{S}\setminus\left(\mathbb{S}_{q_0}\bigcup\mathbb{S}_{q_1}\right),\text{ and } p^*(\mathbf{s})>0,\forall \mathbf{s}\in\mathbb{S}_{q_0}\bigcup\mathbb{S}_{q_1},\nonumber\\
     \mathbb{S}_{q_0}&=\left\{\mathbf{s}_{q_0} \left\rvert P\left(Z=0\mid\mathbf{s}_{q_0}\right)=\max_{\mathbf{s}\in\mathbb{S}} P\left(Z=0\mid \mathbf{s}\right)\right\}\right.,\nonumber\\
     \mathbb{S}_{q_1}&=\left\{\mathbf{s}_{q_1}\left\rvert P\left(Z=1\mid \mathbf{s}_{q_1}\right)=\max_{\mathbf{s}\in\mathbb{S}} P\left(Z=1\mid \mathbf{s}\right)\right\}\right..\label{convergedP}
 \end{align} 
\label{Bimodal-type}
\end{definition}

\label{sec:bimodal}
\section{A batch label-efficient two-sample test}
\label{sec:batch_activetest}
In this section, we propose a three-stage framework for a batch label-efficient two-sample test and outline its statistical properties. We refer interested readers to \shortciteNP{li2022label} for a more comprehensive presentation.
\subsection{A Three-Stage Two-Sample Testing Framework}
The algorithmic description of the three-stage testing framework is listed in Algorithm~\ref{Twosamplealgo}. The inputs of the algorithm~\ref{Twosamplealgo} are as follows: an unlabeled feature set $\mathcal{S}_u$, a classification algorithm $\mathcal{A}$ that takes a training set as input and outputs a classifier, the number $N_0$ of labels used to construct a training set, the label budget $N_q$ and a pre-defined significance level $\alpha$. 
The output of algorithm~\ref{Twosamplealgo} is a single bit of information: was the null hypothesis $H_0$ rejected? During the \textbf{first stage}, a classification algorithm $\mathcal{A}$ takes $N_0$ uniformly labeled samples (and corresponding labels provided by the oracle) as a training set input, and outputs a class-probability estimation function $Q\left(z\mid \mathbf{s}\right)$ used to model $P\left(z|\mathbf{s}\right)$ subsequently. During the \textbf{second stage}, we propose a bimodal query algorithm that queries the labels of samples with highest class one probability $Q\left(1\mid\mathbf{s}\right)$ and highest class zero probability $Q\left(0\mid\mathbf{s}\right)$ until the label query budget, $N_q$, is exhausted.  During the \textbf{third stage}, we split a labeled feature set $\overline{\mathcal{S}}$ to two samples $\overline{\mathcal{X}}$ and $\overline{\mathcal{Y}}$, where each set only contains features from one class. Then the FR two-sample test is performed with the following steps: (1) compute the FR statistic (see Section~\ref{subsubsec:FRTest}) from $\overline{\mathcal{X}}$ and $\overline{\mathcal{Y}}$; (2) compute $p$-value; (3) rejects the null hypothesis if the $p$-value is smaller than the pre-defined significance level $\alpha$. 
\begin{algorithm}[h!]
 \algorithmicinput\hspace{0.2cm}$\mathcal{S},N_0,N_q,\alpha, \mathcal{A}$\\
\algorithmicoutput\hspace{0.2cm} Reject or accept $H_0$\vspace{0.1cm}\\
\algorithmicinit\\ 
Uniformly sample $N_0$ features $\overline{\mathcal{S}}\subset \mathcal{S}_u$ and query their labels $\overline{\mathcal{Z}}$; $\mathcal{S}_u=\mathcal{S}_u/\overline{\mathcal{S}}$;\\
$\mathcal{A}$ takes input $\overline{\mathcal{S}}$ and $\overline{\mathcal{Z}}$, and outputs class-probability predictor $Q(z\mid\mathbf{s})$ used to model $P(z\mid \mathbf{s})$;\vspace{0.1cm}\\
\algorithmicquery\\
Select $\lfloor(N_q - N_0)/2\rfloor$ features $\overline{\mathcal{S}}_0\subseteq\mathcal{S}_u$ which corresponds to $\lfloor(N_q - N_0)/2\rfloor$ highest $Q(0\mid\mathbf{s})$, and query their labels $\overline{\mathcal{Z}}_0$;\\
Select $N_q-N_0 - \lfloor(N_q - N_0)/2\rfloor$ features $\overline{\mathcal{S}}_{1}\subseteq\mathcal{S}_u$ which corresponds to $N_q-N_t - \lfloor(N_q - N_0)/2\rfloor$ highest $Q(1\mid\mathbf{s})$, and query their labels $\overline{\mathcal{Z}}_1$;\\
$\overline{\mathcal{S}}=\overline{\mathcal{S}}\bigcup \overline{\mathcal{S}}_0\bigcup\overline{\mathcal{S}}_1$;
$\overline{\mathcal{Z}}=\overline{\mathcal{Z}}\bigcup \overline{\mathcal{Z}}_0\bigcup\overline{\mathcal{Z}}_1$
\\
\algorithmictest\\
Split $\overline{\mathcal{S}}$ to two samples $\overline{\mathcal{X}}$ and $\overline{\mathcal{Y}}$ based on the label set $\overline{\mathcal{Z}}$;
compute FR statistic using  $\overline{\mathcal{X}}$ and $\overline{\mathcal{Y}}$; compute $p$-value;\\
\textbf{If} $p<\alpha$ \textbf{Then} Reject $H_0$ \textbf{Else} Accept $H_0$.
 \caption{A three-stage framework for the batch label-efficient two-sample testing}
 \label{Twosamplealgo}
\end{algorithm}
\subsection{Consistent Bimodal Query Minimizes the FR Statistic $W_n$}
\vspace{-0.2cm}
\label{SecOpt}
Our problem statement in Section~\ref{sec:activetest} assumes that 
the original unlabeled feature set $\mathcal{S}_u$ includes i.i.d realizations of $\mathbf{S}\sim p\left(\mathbf{s}\right)$, and that the access to every $\mathbf{s}_i\in\mathcal{S}$ is free; but it is costly to obtain the corresponding label $z_i\in\mathcal{Z}$. However, we are assigned a label budget $N_q$ such that we can select a set $\overline{\mathcal{S}}\subseteq\mathcal{S}$ to query labels from an oracle, and each random variable $Z_i$ corresponding to the returned label $z_i$ admits $P(z|\mathbf{s}_i)$. We then divide $\overline{\mathcal{S}}$ to $\overline{\mathcal{X}}$ from class zero and $\overline{\mathcal{Y}}$ from class one and perform a two-sample test on $\overline{\mathcal{X}}$ and $\overline{\mathcal{Y}}$. 
We write $|\overline{\mathcal{X}}|=\overline{n}_0$ and $|\overline{\mathcal{Y}}|=\overline{n}_1$ and we have $N_q=\overline{n}_0+\overline{n}_1$. 

Our aim is to find a query scheme that increases the testing power of a test performed on the selected samples $\overline{\mathcal{X}}$ and $\overline{\mathcal{Y}}$. For a uniform sampling query scheme, then we will have $\overline{\mathcal{S}}$ as a set of $N_{q}$ \textit{i.i.d} realizations generated from the original marginal distribution $p(\mathbf{s})$, and we can rewrite $p$-value in~\eqref{Pvalue} as $p = \Phi[W_{N_q}]$ where $W_{N_q}$ is a FR statistic random variable obtained from $N_q$ \textit{i.i.d.} pairs of $(\mathbf{S}_i, Z_i)\sim p(\mathbf{s},z)$. Instead of directly tackling the query scheme, 
we consider to find an optimal marginal distribution $p_q(\mathbf{s})$ such that, under $H_1$,  performing the FR test on a set of \textit{i.i.d.} $\mathbf{S}\sim p_q(\mathbf{s})$ generates large testing power  than performing on the uniformly sampled data points with the same number of labels $N_q$. 

Under $H_0$, the feature variable $\mathbf{S}$ is always independent of label variable $\mathbf{Z}$ and the Type I error can be controlled by running a permutation test or approximation of that regardless of $p_q\left(\mathbf{s}\right)$. Here, we focus on $p_q\left(\mathbf{s}\right)$ under $H_1$. Specifically, given $N_q$ i.i.d. realizations generated from $p_q\left(\mathbf{s}\right)$, 
we seek a $p_q\left(\mathbf{s}\right)$ to minimize the FR statistic $W_{N_q}$ in~\eqref{FRstat}, thereby creating a more powerful FR test under $H_1$. Before presenting our theoretical analysis, we need the following proposition, which follows from Theorem~\ref{TheoremHenze}. 
\begin{proposition}
    Suppose $\left(\mathbf{S}, Z\right)^n$ comprises \textit{i.i.d.} pairs $\{\mathbf{S}, Z\}\sim p\left(\mathbf{s}, z\right)$, and let $u=P(Z=0)$ and $v=P(Z=1)$. Then, the following holds for a normalized FR statistic $\frac{W_n}{n}$ generated from $\left(\mathbf{S}, Z\right)^n$,  
\begin{align}
  \frac{W_n}{n}\inas \frac{[\int 2 P(Z=0\mid \mathbf{s})p(Z=1\mid \mathbf{s})p(\mathbf{s})d\mathbf{s} - 2uv]}{\sqrt{2uv[2uv + (A_d -1)(1-4uv)]}}
  \label{HP}
\end{align}
where $A_d$ is a constant dependent on the dimension $d$ of $\mathbf{S}$.  
\label{HP2}
\end{proposition}
We direct readers to Section 7 in~\shortciteNP{li2022label} for the proof.
From Proposition~\ref{HP2} we know that the convergence of $\frac{W_{N_q}}{N_q}$ only depends on $p(\mathbf{s})$. Therefore, we construct the following optimization problem: 
\begin{align}
    &\min_{p_q(\mathbf{s})}\int P(Z=0\mid \mathbf{s})P(Z=1\mid \mathbf{s})p_q(\mathbf{s})d\mathbf{s}\nonumber\\
    &\text{subject to}\int P(Z=0\mid \mathbf{s})p_q(\mathbf{s})d\mathbf{s}=u,\quad \int p_q(\mathbf{s})d\mathbf{s}=1,\quad p_q(\mathbf{s})\geq 0.\label{OptFR}
\end{align} 
Under the null hypothesis $H_0$, $Z$ and $S$ are independent and thus $p(\mathbf{s},z) = p(\mathbf{s})p(z)$, and $\int P(Z=0|\mathbf{s})P(Z=1|\mathbf{s})p_q(\mathbf{s})ds=uv$ for any $p_q(\mathbf{s})$. Therefore, minimizing~\eqref{OptFR} with $p_q(\mathbf{s})$ does not alter the Type I error. On the other hand, under the alternate $H_1$,  solving the optimization problem~\eqref{OptFR} leads to a solution that minimizes $W_{N_q}$ in~\eqref{Pvalue} for large sample sizes $N_q\to\infty$, leading to a decreasing Type II error of the FR test.
\begin{theorem}
The optimal solution to~\eqref{OptFR} is $p^*(\mathbf{s})$ defined in~\eqref{convergedP}.
\label{theolinopt}
\end{theorem}
We direct readers to Section 8 in~\shortciteNP{li2022label} for the proof. Briefly, the proof of Theorem~\ref{theolinopt} comes about when we combine the linear constraints in Eq.~\eqref{OptFR} with the fact that the optimum value is always achieved on the boundary of the constraint set for Linear Programming (LP)~\shortcite{korte2011combinatorial}.  The optimal solution $p^*(\mathbf{s})$ of~\eqref{OptFR} is a pdf that samples the highest posterior probabilities of $P(Z=0|\mathbf{s})$ and $P(Z=1|\mathbf{s})$.  \looseness=-1
\begin{remark}
Let $\mathbb{S}$ be the support of $p(\mathbf{s})$ and let $\mathcal{S}_u$ be an unlabeled set where sample features are \textit{i.i.d.} sampled from $p(\mathbf{s})$. Suppose $p_n(\mathbf{s})$ denotes the pdf of $\mathbf{S}_n$, whose label $Z_n$ is queried by the \textit{consistent bimodal query} (see Definition~\ref{Bimodal-type}) for every $n > 0$. Then $p_n(\mathbf{s})$ converges to the optimal solution of~\eqref{OptFR}. In the finite-sample regime, we use a bimodal query where the class-probability predictor $Q(z \mid \mathbf{s})$ is an approximation of $P(z \mid \mathbf{s})$ to query the labels of features in the regions with high posterior probability for class one and zero.
\end{remark}
\subsection{Type I Error of the Three-Stage Framework}
\label{TypeI}
One important observation for the proposed framework is that the features labeled in the second stage are dependent on the uniform sampled features in the first stage. For every $n$ i.i.d. realizations $(\mathbf{s}, z)^n$ of $\left(\mathbf{S}_i, Z_i\right)^n\sim p(\mathbf{s},z)$ under the null hypothesis $H_0$, we write $\overline{\mathcal{S}}=\{\bar{\mathbf{s}}_1,\ldots, \bar{\mathbf{s}}_{N_q}\}$ to denote a set that our query scheme (comprised of uniform sampling and bimodal query) selects from the original unlabelled set $\mathcal{S}_u$, and write $\overline{\mathcal{Z}}=\{\bar{z}_1,\ldots,\bar{z}_{N_q}\}$ to denote a set of label observations corresponding to $\overline{\mathcal{S}}$. We use $\bar{\mathbf{S}}_i$ and $\bar{Z}_i$ to denote the random variables corresponding to $\bar{\mathbf{s}}_i$ and $\bar{z}_i$.  In the following, we present our theorem regarding the Type I error control: 
\begin{theorem}
Suppose $\left(\bar{\mathbf{S}}, \bar{Z}\right)^{N_q}$ are pairs of random feature variables and label variables acquired in the end of the second stage of the framework, using a permutation test in the third stage of the framework to obtain $p$-value from $\left(\bar{\mathbf{S}}, \bar{Z}\right)^{N_q}$ for any two-sample test have Type I error  $P(p\leq\alpha)\leq\alpha,\forall \alpha$ for the framework.
\label{TypeITheory}
\end{theorem}
We direct readers to Section 10 in~\shortciteNP{li2022label} for a detailed proof. Theorem~\ref{TypeITheory} states that the Type I error of our framework is upper-bounded by $\alpha$ for any two-sample test carried out as a permutation test in the third stage. A permutation test rearranges labels of features, obtains permutation distribution of a statistic computed from the rearrangements, and rejects $H_0$ if a true observed statistic is contained in $\alpha$ probability range of the permutation distribution. This process does not need features to be \textit{i.i.d.} sampled to control the Type I error at exact $\alpha$,  and it is applicable to any two-sample tests testing independency between $\bar{\mathbf{S}}_i$ and $\bar{Z}_i$. However, 
we need to make sure our query procedure maintains $\bar{\mathbf{S}}_i \indep \bar{Z}_i$ under the $H_0$.
Our framework only trains a classifier one time with uniformly sampled data points in the first stage, and then the bimodal query selects a subset of features from $\mathcal{S}$ to label based on the trained classifier. For a set of feature and label variables $\mathcal{Q}=\{\bar{\mathbf{S}}_i, \bar{Z}_i\}_{i=1}^{N_q}$ obtained in the end of the second stage, we write  $\mathcal{Q}_{u}\subseteq\mathcal{Q}$ to denote the set obtained from uniform sampling, and write $\mathcal{Q}_{b}\subseteq\mathcal{Q}$ to denote the set  obtained from bimodal query. Considering that a uniform sampling scheme does not change the original distributional properties ($\mathbf{S} \indep Z$ under the null) to generate $(\bar{\mathbf{S}}_i, \bar{Z}_i)\in\mathcal{Q}_u$, we have $\bar{\mathbf{S}}_i \indep \bar{Z}_i, \forall (\bar{\mathbf{S}}_i, \bar{Z}_i)\in\mathcal{Q}_u$. $\mathcal{Q}_{b}$ is not used to train the classifier, so we also have $\bar{\mathbf{S}}_i \indep \bar{Z}_i,\forall (\bar{\mathbf{S}}_i, \bar{Z}_i)\in\mathcal{Q}_b$. 

\section{A Sequential Label-Efficient Two-sample Test}
\label{sec:sequential_activetest}

The batch design discussed in Section~\ref{sec:batch_activetest} invariably exhausts the label budget. Fixing the number of labels  can lead to inefficiencies: if the two-sample testing problem is difficult, additional label query for more evidence may be required for a final decision, whereas if the problem is simple, the test may collect excessive labels. Therefore, this section introduces a sequential label-efficient two-sample testing framework that adapts the number of label queries based on the problem's complexity while maintaining statistical validity. For a more comprehensive presentation, we refer interested readers to \shortciteNP{li2023active}.

\subsection{A Sequential Label-Efficient Framework}
\begin{figure}[H]
\begin{tcolorbox}
Suppose $N_q$ is a label budget and $\alpha$ is a significance level.  An analyst uses the proposed framework to sequentially and actively query the label $z_{n}$ of $\mathbf{s}_{n}$ from an unlabelled feature set $\mathcal{S}_u$ based on the predictions of $Q_{n}\left(z\mid \mathbf{s}\right)$. As a new $z_n$ of $\mathbf{s}_n$ is queried, the analyst constructs the following statistic, 
\begin{align}
u_{n}=\prod_{i=1}^{n}\frac{\hat{P}(z_i)}{Q_{i}\left(z_i\mid \mathbf{s}_i\right)}.
\label{ratiostatisics2}
\end{align}
The analyst evaluates $w_n$ and makes one of the following decisions: (1) rejects $H_0$ if $u_{n}\leq\alpha$; (2) retains $H_0$ if the label budget $N_q$ is exhausted and (1) is not satisfied; and (3) continues the test and updates $Q_{n}$ to $Q_{n+1}$ if (1) and (2) are not satisfied.
\end{tcolorbox}
\caption{The sequential label-efficient framework}
\label{ActiveSequentialFramework}
\end{figure}
In~\shortciteNP{li2023active}, we introduced a novel framework for the sequential label-efficient two-sample test.  We use a statistic $u_n$ slightly modified from the sequential testing statistic $v_n$~\eqref{ratiostatisics}. Given that labels are costly to access in the problem setting (See Section~\ref{sec:activetest}), the class prior $P(z)$ required to compute~\eqref{ratiostatisics} is also unknown.  Therefore, \textit{ we will use a likelihood estimate $\hat{P}(z^n)$, maximized over all class priors, to replace $P(z^n)$—the product of the class prior in~\eqref{ratiostatisics}}. This results in $u_n=\frac{\hat{P}(z^n)}{Q(z^n\mid \mathbf{s}^n)}=\prod_{i=1}^n\frac{\hat{P}(z_i)}{Q_{i}(z_i\mid \mathbf{s}_i)}$ where $\hat{P}(Z=1)=\frac{\sum_{i=1}^{n}z_i}{n}$. Our framework begins by labeling sample features randomly selected from from $\mathcal{S}_u$, the unlabeled feature set, and then initializing the class-probability predictor $Q_{n}(z\mid 
\mathbf{s})$ at $n=1$ with the labeled features. Then, the framework enters the sequential testing stage to sequentially query label, update and monitor the statistic 
$u_n$. We formally introduce our framework in Fig~\ref{ActiveSequentialFramework}. 

We provide a \textit{framework instantiation} called \textit{bimodal query based active sequential two-sample testing (BQ-AST)} described in Algorithm~\ref{instaalgo}. The algorithm takes the following input: an unlabelled feature set $\mathcal{S}_u$, a probabilistic classification algorithm $\mathcal{A}$, the size $N_0$ of an initialization set used for $\mathcal{A}$, a label budget $N_q$ and a significance level $\alpha$. Then, the algorithm initializes a class-probability predictor $Q$ using $\mathcal{A}$ with a small set of randomly labeled samples. In the sequential testing stage, the algorithm uses bimodal query discussed in Section~\ref{Sec:Bimodal-query} to sample $\mathbf{s}_n$ with samples having the highest posteriors from either class (e.g. a fair chance to select the highest $Q_{n}\left(Z=0\mid \mathbf{s}\right)$ or $Q_{n}\left(Z=1\mid \mathbf{s}\right)$) from $\mathcal{S}_u$, queries its label $z_n$ and updates the statistic $w_n$. Next, the algorithm compares $w_n$ with $\alpha$ and make a decision, e.g., reject $H_0$ or continue testing. \textit{In what follows in this section, we simply use $N_q$ to denote the ``label budget'' allowed to be used after the initialization}.\looseness=-1

\begin{minipage}{0.95\textwidth}
\centering
\begin{algorithm}[H]
\centering
\caption{Bimodal Query Based Active Sequential Two-Sample Testing (BQ-AST)}
   \label{instaalgo}
\begin{algorithmic}[1]
\State{\bfseries Input:} $\mathcal{S}_u, \mathcal{A}, N_0, N_q, \alpha$
\State{\bfseries Output:} Reject or fail to reject $H_0$
\State\textbf{Initialization}: Initialize $Q_1(z\mid \mathbf{s})$ using $\mathcal{A}$ with $N_0$ features uniformly sampled from $\mathcal{S}_u$ without replacement and then labeled. 
\State\textbf{Active Sequential testing}: 
\For{$n = 1$ to $N_q-N_0$}
\State Sample a feature $\mathbf{s}_{n}=\mathbf{s}_{q_0}$ or $\mathbf{s}_{q_1}$ with \textbf{fair chance} where $\mathbf{s}_{q_0}=\arg\max_{\mathbf{s}}\left[Q_{n}(Z=0|\mathbf{s})\right],\forall \mathbf{s}\in\mathcal{S}_u$ and $\mathbf{s}_{q_1}=\arg\max_{\mathbf{s}}\left[Q_{n}(Z=1|\mathbf{s})\right],\forall \mathbf{s}\in\mathcal{S}_u$
\State Query the label $z_{n}$ of $\mathbf{s}_{n}$
\State Update $u_n$ in~\eqref{ratiostatisics2} with $(\mathbf{s}_n, z_n)$ and $Q_{n}(z_n\mid \mathbf{s}_n)$
\If{$u_n\leq\alpha$}
\State\textbf{Return} Reject $H_0$
\Else
\State \textbf{Update} $Q_{n}(z\mid \mathbf{s})$  with newly queried $(\mathbf{s}_n, z_n)$ and past training examples. 
\EndIf
\EndFor
\State\textbf{Return} Retain $H_0$
\end{algorithmic}
\end{algorithm}
\end{minipage}
\subsection{The Proposed Framework Results in an Anytime-Valid $p$-value}
\label{SecTypeIControl}
Our framework rejects $H_0$ if the statistic $u_{n}\leq\alpha$. The following theorem states that under $H_0$, $u_{n}$ is an \emph{anytime-valid} $p$-value.  
\begin{theorem}
 If an analyst uses the proposed framework to sequentially query the oracle for $Z$ with $\mathbf{S}\in\mathcal{S}_u$ resulting in $(\mathbf{S},Z)^n$, then we have the following under $H_0$, 
\begin{align}
P_0\left(\exists n\in \left[N_q\right], U_{n}=\prod_{i=1}^{n}\frac{\hat{P}(Z_i)}{Q_{i}\left(Z_i\mid \mathbf{S}_i\right)}\leq\alpha\right)\leq\alpha\nonumber
\end{align}
where $N_q$ is a label budget and $\alpha$ is the pre-specified significance level. 
\label{ActiveTypeI}
\end{theorem}
We direct readers to Section A in~\shortciteNP{li2023active} for the detailed proof. Theorem~\ref{ActiveTypeI} implies the probability $P_0$ (or Type I error) that our framework mistakenly rejects $H_0$ is upper-bounded by $\alpha$. The proof of Theorem~\ref{ActiveTypeI} builds upon Theorem~\ref{PassiveTypeI}, and we can demonstrate that $P(U_n\geq T_n)=1,\forall n$, assuming  $t_n$~\eqref{ratiostatisics} is constructed from the same $(s,z)^n$ with a known class prior $P(z)$. 
\subsection{Asymptotic Properties of the Proposed Framework}
\label{AsympoticProp}
We consider  normalizing the test statistic in~\eqref{ratiostatisics2} as follows, 
\begin{align}
    \overline{U}_{n}=\frac{1}{n}\sum_{i=1}^{n}\log\frac{\hat{P}(Z_i)}{Q_{i}\left(Z_i\mid \mathbf{S}_i\right)},\left(\mathbf{S}_i, Z_i\right)\sim p_{i}\left(\mathbf{s},z)=p(z\mid \mathbf{s}\right)p_{i}\left(\mathbf{s}\right)\nonumber
\end{align}
where $(\mathbf{S}_i, Z_i)$ denotes a feature-label pair returned by a label query scheme when querying the $i$-th label. We then state the following theorem.

\begin{theorem}
 Let $\mathbb{S}$ be the support of  $p\left(\mathbf{s}\right)$ that sample features 
 are collected from and added to an unlabeled set $\mathcal{S}_u$, and let $P\left(z\mid \mathbf{s}\right)$ denote the posterior probability of $z$ given $\mathbf{s}\in\mathcal{S}$. There exists a consistent bimodal query scheme (see Definition~\ref{Bimodal-type}); when an analyst uses such a scheme in the proposed active sequential framework, then, under $H_1$, $\overline{U}_{n}$ converges to the negation of mutual information (MI), and the converged negated MI lower-bounds the negated MI generated by any $p\left(\mathbf{s}\right)$ subject to $P\left(z\mid\mathbf{s}\right),\forall \mathbf{s}\in\mathbb{S}$. Precisely, there exists a consistent bimodal query leading to the following
\begin{align}
\lim_{n\to\infty}\overline{U}_{n}= -\left(H^*(Z)-H^*\left(Z\mid \mathbf{S}\right)\right)=-I^*\left(\mathbf{S}; Z\right)\leq -I\left(\mathbf{S}; Z\right)\nonumber.
\end{align}
$I^*\left(\mathbf{S};Z\right)$ is the MI constructed with $\left(\mathbf{S}, Z\right)\sim p^*\left(\mathbf{s}, z\right)=P\left(z\mid \mathbf{s}\right)p^*\left(\mathbf{s}\right)$ (See~\eqref{convergedP} for $p^*\left(\mathbf{s}\right)$); $I\left(\mathbf{S}; Z\right)$ is MI constructed with $\left(\mathbf{S}, Z\right)\sim p\left(\mathbf{s}, z\right)=P\left(z\mid \mathbf{s}\right)p\left(\mathbf{s}\right)$. 
\label{TheoryAsym}
\end{theorem}
We direct readers to Section B in~\shortciteNP{li2023active} for the detailed proof. Recalling the null $H_0$ is rejected when the test statistic $u_n$ in ~\eqref{ratiostatisics2} is smaller than $\alpha$; hence, the proposed framework, when used with a consistent bimodal query to asymptotically minimize the normalized  $u_n$ in ~\eqref{ratiostatisics2}, favorably increases the testing power when $|\mathcal{S}_u|$ is large and $Q(z\mid\mathbf{s})$ is close to $P(z\mid\mathbf{s})$. Theorem~\ref{TheoryAsym} alludes that the proposed framework asymptotically turns the original hard two-sample testing problem to a simply by increasing the dependency between $\mathbf{S}$ and $Z$.  
\begin{remark}
Our testing framework is also \textit{consistent} under $H_1$ and the same conditions of Theorem~\ref{TheoryAsym} as $\lim_{n\to\infty}P_1\left(\prod_{i=1}^{n}\frac{\hat{P}(Z_i)}{Q_{i}(Z_i\mid \mathbf{S}_i)}\leq\alpha\right)=\lim_{n\to\infty}P\left(\overline{U}_{n}\leq\frac{1}{n}\log(\alpha)\right)=P_1(-I^*\left(\mathbf{S}, Z\right)\leq 0)=1$. The last equality holds due to $I^*\left(\mathbf{S}, Z\right)>0$ under $H_1$. 
\end{remark}

\subsection{Finite-Sample Analysis for the Proposed Framework}
\label{seqfinite}
This section analyzes the testing power of the proposed framework in the finite-sample case. Section~\ref{SecApproxErr} and~\ref{SecIrreducibleErr} offer metrics that assess the approximation error of $Q(z\mid\mathbf{s})$ and an irreducible Type II error. They together determine the finite-sample testing power. Furthermore, Section~\ref{SecExample} presents an illustrative example of using our framework. In Section~\ref{SecFiniteAnlysis}, we conduct a finite-sample analysis for the example, incorporating both the metrics that characterize the approximation error and the irreducible Type II error.
\subsubsection{Characterizing the Approximation Error of $Q(z\mid\mathbf{s})$}
\label{SecApproxErr}
As our framework constructs the test statistic in~\eqref{ratiostatisics} with the approximation $Q(z\mid\mathbf{s})$, there arises a need to establish a metric for assessing the approximation error of $Q(z\mid\mathbf{s})$ for our finite-sample analysis. To this end,  we introduce $\text{KL}^{2}$-divergence,
\begin{definition}{(\textbf{$\text{KL}^{2}$-divergence})}
    Let $p_0$ and $q_0$ be two probability density functions on the same support $\mathcal{X}$. Let $f(t)=\log^2(t)$. Then, the $\text{KL}^{2}$-divergence between $p_0$ and $q_0$ is 
    \begin{align}
        D_{\text{KL}^2}\left(q_0\|p_0\right) = \mathbb{E}_{\mathbf{X}\sim p_0(\mathbf{x})}\left[f\left(\frac{q_0(\mathbf{X})}{p_0(\mathbf{X})}\right)\right]=\mathbb{E}_{\mathbf{X}\sim p_0(\mathbf{x})}\left[\log^2{\left(\frac{q_0(\mathbf{X})}{p_0(\mathbf{X})}\right)}\right]\nonumber.
    \end{align}
\end{definition}
$D_{\text{KL}^2}\left(q_0\|p_0\right)$ is the second moment of the log-likelihood ratio and has been used (see, e.g., (3.1.14) in~\shortciteNP{koga2002information}) to understand the behavior of the distribution of $\log{\left(\frac{q_0(\mathbf{x})}{p_0(\mathbf{x})}\right)}$. We use $D_{\text{KL}^2}\left(q_0|| p_0\right)$ to evaluate the distance between  $p\left(\mathbf{s}, z\right)=P\left(z\mid \mathbf{s}\right)p\left(\mathbf{s}\right)$ and $q(\mathbf{s}, z)=Q\left(z\mid \mathbf{s}\right)p\left(\mathbf{s}\right)$, which 
yields the following 
\begin{align}
  D_{\text{KL}^2}\left(q(\mathbf{s}, z)\| p(\mathbf{s}, z)\right)=\mathbb{E}_{\left(\mathbf{S}, Z\right)\sim p(\mathbf{s}, z)}\left[\log^2\left(\frac{q\left(\mathbf{S}, Z\right)}{p\left(\mathbf{S}, Z\right)}\right)\right]=\mathbb{E}_{\left(\mathbf{S}, Z\right)\sim p(\mathbf{s}, z)}\left[\log^2\left(\frac{Q\left(Z\mid\mathbf{S}\right)}{P\left(Z\mid\mathbf{S}\right)}\right)\right].
\label{KLSquarePQ1}
\end{align}
Remarkably, $D_{\text{KL}^2}\left(q(\mathbf{s}, z)\|p(\mathbf{s}, z)\right)$ in~\eqref{KLSquarePQ1} also characterizes the discrepancy between $P\left(z\mid\mathbf{s}\right)$ and $Q\left(z\mid\mathbf{s}\right)$ by averaging their log square distance over $\mathcal{S}$. \textit{$D_{\text{KL}^2}\left(q\left(\mathbf{s}, z\right)\| p\left(\mathbf{s}, z\right)\right)$ and $D_{\text{KL}^2}\left(p\left(\mathbf{s}, z\right)\| q\left(\mathbf{s}, z\right)\right)$ both characterize the approximation error of $Q(z\mid\mathbf{s})$, and in our main results, we will also see they jointly determine the testing power of the proposed framework in Section~\ref{SecFiniteAnlysis}}.

\subsubsection{Characterizing the Factor that Leads to the Irreducible Type II Error in Finite-Sample Case}
\label{SecIrreducibleErr}
We also introduce another factor influencing testing power, which persists even in the absence of approximation error, i.e., $Q(z\mid\mathbf{s})=P(z\mid\mathbf{s})$. To see this, we recall the following random variable,
\begin{align}
    \bar{I}(\mathbf{S}^n;Z^n)=\frac{1}{n}\sum_{i=1}^n\log\frac{p\left(\mathbf{S}_i, Z_i\right)}{p_{\mathbf{S}}(Z_i)p_{Z}(Z_i)},\quad \left(\mathbf{S}, Z\right)\sim p(\mathbf{s}, z)
\nonumber
\end{align}
The distribution of $\bar{I}$ is formally defined as the \textit{information spectrum} in~\shortciteNP{han1993approximation}.~\shortciteNP{han2000hypothesis} leverages the dispersion of $\bar{I}(\mathbf{S}^n;Z^n)$  to quantify the rate that Type II error goes to zero with increasing $n$. Their underlying rationale is that, for a larger variance of $\bar{I}$, the probability of $\bar{I}$ falling outside the acceptance region for an alternative hypothesis also increases, thereby resulting in a slower convergence rate for the Type II error. In our work, we will make use of the variance of the log-likelihood ratio between $p(\mathbf{s}, z)$ and  $p(\mathbf{s})p(z)$ \looseness=-1
\begin{align}
    \text{Var}_{(\mathbf{S}; Z)\sim p(\mathbf{s}, z)}\bar{I}(\mathbf{S}, Z)=n\text{Var}_{(\mathbf{S}; Z)^n\sim p\left(\left(\mathbf{s}, z\right)^n\right)}\bar{I}(\mathbf{S}^n, Z^n)=\text{Var}_{\left(\mathbf{S},Z\right)\sim p\left(\mathbf{s},z\right)}\left[-\log\frac{P(Z)}{P\left(Z\mid\mathbf{S}\right)}\right]\nonumber.
\end{align}
It remains present even in the absence of approximation error (i.e., $Q(z\mid\mathbf{s})=P(z\mid\mathbf{s})$). \textit{As we will see in Section~\ref{SecFiniteAnlysis}, the persistent $\text{Var}_{(\mathbf{S}, Z)\sim p(\mathbf{s}, z)}\bar{I}(\mathbf{S}; Z)$ leads to a non-zero Type II error in the finite-sample case.}
\begin{figure}[H]
\begin{tcolorbox}
(\textit{\textbf{An example of the proposed framework}}) Given a label budget $N_q$, $\alpha$, an unlabeled set $\mathcal{S}_u$, a partition $\mathcal{P}=\{A_1,\cdots, A_m\}$, and class priors $\left\{P(Z=0\mid A_1),\cdots, P(Z=0\mid A_m)\right\}$, an analyst initializes $Q\left(z\mid \mathbf{s}\right)$ with a set of labeled features randomly sampled from $\mathcal{S}_u$, then, she estimates $I\left(\mathbf{S}; Z\mid A_i\right)$ by 
\begin{align}
    \hat{I}&\left(\mathbf{S}; Z\mid A_i\right) = H\left(Z\mid A_i\right) - \hat{H}\left(Z\mid\mathbf{S}, A_i\right)\nonumber\\
    &=-\sum_{z=0}^1 P\left(Z=z\mid A_i\right) \log P\left(Z=z\mid A_i\right) + \frac{\sum_{\mathbf{s}\in A_i\bigcap \mathcal{S}_u}\sum_{z=0}^1 Q\left(Z=z\mid \mathbf{s}\right) \log Q\left(Z=z\mid \mathbf{s}\right)}{\left|A_i\bigcap \mathcal{S}_u\right|}\label{MIEstimate},
\end{align}
selects $A^*=\arg\max_{A\in\mathbb{P}}\hat{I}\left(\mathbf{S};Z\mid A\right)$, and sequentially constructs the statistic $u_n=\prod_{i=1}^n \frac{P(z_i)}{Q\left(z_i\mid \mathbf{s}_i\right)}$ by labelling features randomly sampled from $A^*\bigcap\mathcal{S}_u$. The analyst rejects $H_0$ whenever $u_n\leq\alpha$ or retains $H_0$ if the label budget runs out.\\
(\textit{\textbf{The baseline}}) Given a label budget $N_q$, $\alpha$, an unlabeled set $\mathcal{S}_u$ and the class prior $P(Z=0)$, an analyst initializes $Q(z\mid \mathbf{s})$ with a set of labeled features randomly sampled from $\mathcal{S}_u$, then, she sequentially constructs the statistic $u_n=\prod_{i=1}^n \frac{P(z_i)}{Q\left(z_i\mid \mathbf{s}_i\right)}$ by labelling features randomly sampled from $\mathcal{S}_u$. The analyst rejects $H_0$ whenever $u_n\leq\alpha$ or retains $H_0$ if the label budget runs out. 
\end{tcolorbox}
\caption{An example of the proposed framework and the baseline.}
\label{FrameworkExample}
\end{figure}
\subsubsection{An Example of Using the Proposed Framework}
\label{SecExample}
 We write $\mathcal{P}=\left\{A_1,\cdots, A_m\right\}$ to denote a partition of the support $\mathbb{S}$ of $p(\mathbf{s})$ from which unlabeled sample features in $\mathcal{S}_u$ are generated; in other words, $\bigcup_{i=1}^m A_{i}=\mathbb{S}$. In the following, we quantitatively compare an example of the framework to a baseline where features are randomly sampled from $\mathcal{S}_u$ and labeled. The example of the framework and the baseline are as in Fig~\ref{FrameworkExample}.
The class priors $\left\{P\left(z\mid A_i\right)\right\}$ are given to simplify our analytical results. In addition, the label-efficient test chooses the partition $A^*$ predicted by $Q\left(z\mid\mathbf{s}\right)$ to have the highest dependency between $\mathbf{S}$ and $Z$ and only conducts sequential testing with the labeled points in $A^*$. In contrast, the baseline case conducts the sequential test entirely the same, except that the analyst queries the labels of features that are randomly generated from $\mathcal{S}_u$. Both the proposed framework and the baseline cases assert the use of a stable $Q\left(z\mid\mathbf{s}\right)$ with no updates in the sequential testing, which is sufficient and convenient for our analysis.
\subsubsection{Finite-Sample Analysis for the Example}
\label{SecFiniteAnlysis}
We use $ \epsilon_1=\max_{A\in\mathcal{P}}D_{\text{KL}^2}\left(q\left(\mathbf{s}, z\right)\| p\left(\mathbf{s}, z\right)\mid A \right)$ and $ \epsilon_2=\max_{A\in\mathcal{P}}D_{\text{KL}^2}\left(p\left(\mathbf{s}, z\right)\| q\left(\mathbf{s}, z\right)\mid A \right)$ to capture the maximum approximation error of $Q(z\mid \mathbf{s})$ over the partition  $\mathcal{P}=\{A_1,\cdots, A_m\}$, and use $\sigma^2 = \max\left\{\max_{A\in\mathcal{P}}\text{Var}_{\left(\mathbf{S},Z\right)\sim p\left(\mathbf{s},z\mid A\right)}\bar{I}(\mathbf{S}; Z), \text{Var}_{\left(\mathbf{S},Z\right)\sim p\left(\mathbf{s},z\right)}\bar{I}(\mathbf{S}; Z)\right\}$ to capture the maximum irreducible Type II error over the same partition $\mathcal{P}$.  We will need to make the following assumptions.
\begin{assumption}{(\textbf{Maximum mutual 
 information gain})}
$\max_{A\in\mathcal{P}}I\left(\mathbf{S};Z\mid A\right) - I(\mathbf{S};Z)=\Delta \geq 0$.
\label{Assumpa}
\end{assumption}
Assumption~\ref{Assumpa} characterizes the largest MI gain of the proposed framework in the case study over the baseline; that is the direct reason for the increased testing power of the proposed framework.
\begin{assumption}{(\textbf{Sufficient size of unlabeled samples})}
\begin{align}
\frac{\sum_{\mathbf{s}\in A\cap\mathcal{S}_u}\mathbb{E}_{Z\sim Q\left(z\mid\mathbf{s}\right)}\left[\log\left(\frac{Q\left(Z\mid\mathbf{s}\right)}{P\left(Z\mid\mathbf{s}\right)}\right)\right]}{\left|A\cap\mathcal{S}_u\right|}\approx D_\text{KL}\left(Q\left(z\mid\mathbf{s}\right)\|P\left(z\mid\mathbf{s}\right)\mid A\right),\forall A\in\mathcal{P}\nonumber.
\end{align}
\label{Assumpd}
\end{assumption}
Assumption~\ref{Assumpd} assumes a sufficient supply of unlabeled samples to simplify the analysis and concentrate solely on the number of labels needed for the case study of the proposed framework. Now, we present our theorem to address the testing power of the case study in the finite-sample case. 
\begin{theorem}
Under Assumption~\ref{Assumpa} and \ref{Assumpd}, the example of the proposed framework with a label budget $N_q$ and $\alpha$ has a testing power of approximately at least 
\begin{align}
\Phi\left(\frac{\frac{\log\alpha}{\sqrt{N_q}} + \sqrt{N_q}\left(I\left(\mathbf{S};Z\right) + \Delta -2\sqrt{\epsilon}_1 - \sqrt{\epsilon}_2\right)}{\left(\epsilon_1 + \sigma^2 + 2\sigma\sqrt{\epsilon_1}\right)^{1/2}}\right);
\label{FiniteEq1}
\end{align} 
and the baseline test has a testing power of approximately at least 
\begin{align}
\Phi\left(\frac{\frac{\log\alpha}{\sqrt{N_q}} + \sqrt{N_q}\left(I\left(\mathbf{S};Z\right)-\sqrt{\epsilon}_1\right)}{\left(\epsilon_1 + \sigma^2 + 2\sigma\sqrt{\epsilon_1}\right)^{1/2}}\right).
\label{FiniteEq2}
\end{align}
\label{Finite-sample}
\end{theorem}
We direct readers to Section C in~\shortciteNP{li2023active} for the detailed proof. We observe that 
\begin{itemize}[leftmargin=*]
\item Given $\alpha$, the lower bounds of the testing powers for both the proposed framework and the baseline, increase with a larger budget $N_q$ and smaller approximation errors characterized by $\epsilon_1$. 
\item  Comparing~\eqref{FiniteEq1} for the proposed framework to the~\eqref{FiniteEq2} for the baseline, the extra $\Delta$ is ascribed to the maximum power gain,  and $ \sqrt{\epsilon_1}+\sqrt{\epsilon_2}$ accounts for the diminishing of the maximum power gain in selecting a $A^*\in\mathcal{P}$ that does not have the highest MI over $A\in\mathcal{P}$. 
\item When the approximation errors $\epsilon_1=0$ and/or $\epsilon_2=0$,  both testing power's lower-bounds are decreased by a factor of $\sigma$, resulting in the irreducible Type II error. 
\item When the maximum MI gain $\Delta$ can compensate the approximation error of  $Q\left(z\mid \mathbf{s}\right)$ being larger than $\sqrt{\epsilon_1} + \sqrt{\epsilon_2}$, our framework in the example has higher testing power's lower bound than the baseline test given the same label budget $N_q$ and $\alpha$.  
\end{itemize}
\section{Future development}
Beyond the digital health example for Alzheimer’s Disease, there are many other applications where label-efficient two-sample tests improve validation efficiency. In cancer research, genetic biomarkers can be collected non-invasively from patients, but confirming the presence of a specific type of cancer may require invasive biopsies or expensive imaging tests. Using a limited number of biopsy-confirmed cases can help efficiently validate the association between genetic biomarkers and cancer presence. In wildlife studies, data from non-invasive methods like camera traps or acoustic sensors can be collected over large areas, while confirming species presence or health status may necessitate capturing and tagging animals, a costly and labor-intensive process. Similarly, in financial transactions, features such as transaction amount, frequency, and location can be collected automatically, but identifying fraudulent transactions often demands human investigation and verification. Finally, for assessing pollution levels, inexpensive sensor data (e.g., air quality indices) can be collected widely, while confirming the exact pollutant composition and levels may require costly laboratory analyses. In all these examples, label-efficient two sample testing allows researchers to validate sensor data against lab results using a small sample of lab-verified data. 
\section*{Acknowledgement}
This work was funded in part by Office of Naval Research grant N00014-21-1-2615 and by the National Science Foundation (NSF) under grants CNS-2003111, and CCF-2048223.

\footnotesize

\bibliographystyle{wsc}

\bibliography{demobib}
\section*{AUTHOR BIOGRAPHIES}
\noindent {\bf \MakeUppercase{Weizhi Li}} is a postdoctoral associate in the Statistical Science Group at Los Alamos National Laboratory. His research primarily focuses on developing data-efficient machine learning algorithms with statistical guarantees. His email address is \email{weizhili@lanl.gov} and his website is \url{https://wayne0908.github.io/}.\\

\noindent {\bf \MakeUppercase{Visar Berisha}} is a Professor in the College of Engineering and the College of Health Solutions at Arizona State University. At its core, his work is interdisciplinary and use-inspired; it lies at the intersection of engineering and human health and is driven by a desire to use technology to improve the human condition. His email address is \email{visar@asu.edu} and his website is \url{http://vees.ar/}.\\

\noindent {\bf \MakeUppercase{Gautam Dasarathy}} is an Associate Professor in the School of Electrical, Computer and Energy Engineering at Arizona State University. His research interests include various topics in machine learning, high-dimensional statistics, information processing, and networked systems.  His email address is \email{gautamd@asu.edu} and his website is \url{http://gautamdasarathy.com}.
\end{document}